# System Identification Using Kolmogorov-Arnold Networks: A Case Study on Buck Converters

Nart Gashi
*Dept. of Electrical Engineering*
Eindhoven University of Technology
Eindhoven, The Netherlands
n.gashi@student.tue.nl

Panagiotis Kakosimos
*Dept. of Energy Conversion*
ABB AB Corporate Research
Vasteras, Sweden
Panagiotis.Kakosimos@se.abb.com

George Papafotiou
*Dept. of Electrical Engineering*
Eindhoven University of Technology
Eindhoven, The Netherlands
g.papafotiou@tue.nl

*Abstract*— **Kolmogorov-Arnold Networks (KANs) are emerging as a powerful framework for interpretable and efficient system identification in dynamic systems. By leveraging the Kolmogorov-Arnold representation theorem, KANs enable function approximation through learnable activation functions, offering improved scalability, accuracy, and interpretability compared to traditional neural networks. This paper investigates the application of KANs to model and analyze the dynamics of a buck converter system, focusing on state-space parameter estimation along with discovering the system equations. Using simulation data, the methodology involves approximating state derivatives with KANs, constructing interpretable state-space representations, and validating these models through numerical experiments. The results demonstrate the ability of KANs to accurately identify system dynamics, verify model consistency, and detect parameter changes, providing valuable insights into their applicability for system identification in modern industrial systems.**

## I. Introduction

System identification, defined as the modeling and characterization of dynamic systems using measured input-output data, is regarded as a foundational aspect of control engineering, signal processing, and related fields where an understanding of system behavior is essential. Traditionally, this process has been achieved through physics-based modeling grounded in first principles, which has been further enhanced by data-driven techniques, including linear regression and classical statistical methods, to estimate system parameters and structures [1]. These methods have historically enabled significant advancements; however, the increasing complexity, nonlinearity, and data intensity of modern engineering systems have necessitated the development of more versatile and expressive modeling techniques. As an example, robotic manipulators, such as multi-jointed arms employed in industrial automation, exhibit intricate nonlinear behaviors arising from their kinematics and dynamics [2]. The accurate modeling of such systems is required to facilitate the design of robust and precise controllers.

In recent years, system identification has been significantly influenced by advancements in machine learning and deep learning methodologies [1], [3]. These data-driven approaches have demonstrated proficiency in capturing complex nonlinearities and hidden dynamics without necessitating explicit assumptions about the underlying physics [4]. Improvements in accuracy, generalization, and scalability have been achieved through their integration with modern sensor technologies and the extensive datasets generated by complex cyber-physical systems [5]. However, the interpretability of many deep learning models remains a critical challenge [6]. The opaque nature of these models complicates the verification of their correctness, stability, and reliability. Interpretable models are considered essential for providing insights into system dynamics and supporting the design and deployment of controllers [7]. By contrast, the use of black-box deep learning models has been shown to hinder trust, particularly in safety-critical applications where unexpected failures can result in significant consequences.

To address these challenges, hybrid modeling approaches have been developed to incorporate physics-based constraints or known system properties into neural network architectures. Physics-informed neural networks (PINNs) represent one such approach, where domain knowledge is integrated to improve transparency, robustness, and sample efficiency [8]. Additional techniques, including sparse identification of nonlinear dynamical systems (SINDy), have been employed to derive parsimonious and interpretable models by identifying governing equations directly from data [9]. While SINDy does not utilize neural networks, it can complement machine learning techniques by extracting governing equations from data-driven dynamical models. Recent advancements, such as Kolmogorov-Arnold Networks combined with Neural Ordinary Differential Equation theory (KAN-ODEs), have further expanded this area of research by applying neural ODEs to develop interpretable models of dynamical systems [10], [11]. Kolmogorov-Arnold Networks (KANs), in particular, have been shown to provide interpretability in function approximation tasks, making them a suitable framework for modeling dynamic systems.

In this study, the applicability of KANs to system identification is examined, with an emphasis on their capacity to model dynamical systems and validate these models through simulation and computational analysis. Particular attention is given to the use of KANs in key tasks, including system identification and model verification. The methodology involves the application of KANs to analyze simulation data from a buck converter, enabling the derivation of state-space matrices to evaluate interpretability and accuracy. Additionally, the potential of KANs to facilitate the reliable modeling of dynamic systems is highlighted, supporting their suitability for addressing challenges in modern system identification and diagnostics.

## II. METHODOLOGY

This section provides a detailed explanation of the theoretical foundation and development process for KANs. The framework, the network structure, and key mathematical formulations that enable them to approximate the dynamics of complex systems are presented.

### A. Framework of KANs

KANs represent a new class of neural networks grounded in the principles of the Kolmogorov-Arnold representation theorem [10]. This theorem establishes that any multivariate continuous function $f: [0,1]^n \to \mathbb{R}$ can be expressed as:

$$f(\mathbf{x}) = f(x_1, \ldots, x_n) = \sum_{q=1}^{2n+1} \Phi_q \left( \sum_{p=1}^{n} \varphi_{q,p}(x_p) \right), \quad (1)$$

where $\varphi_{q,p}$ and $\Phi_q$ are continuous functions dependent on a single variable. Unlike conventional Multi-Layer Perceptons (MLPs), where nodes employ fixed activation functions, KANs introduce learnable activation functions applied to edges. This architectural change enhances both the flexibility and interpretability of KANs (see Fig. 1).

The KAN architecture gives them an advantage over traditional MLPs in terms of both accuracy and interpretability, especially in function approximation tasks. They benefit from more efficient neural scaling, allowing smaller networks to achieve performance levels comparable to or better than larger MLPs. Furthermore, they are particularly capable of capturing compositional structures and optimizing single-variable functions, which helps alleviate challenges associated with high-dimensional data. Once a KAN network finishes training, symbolic regression techniques can be applied to derive exact equations, from the learned activation functions. Their interpretability is further enhanced through advanced methods such as sparsification, pruning, and symbolic regression. These attributes make them highly suitable for scientific exploration, as they can uncover underlying mathematical and physical principles [10].

### B. Learning dynamic systems

Dynamic systems can be learned using two primary approaches: Simply the "fitting the derivative" method and a more advanced approach which is known as neural ODE method, which places an ODE solver in the training loop. In this work, we adopt the simple "fitting the derivative" approach because it is comparatively less complex. Consider a dynamic system with $n$ states, represented as:

$$\frac{d\mathbf{x}}{dt} = f(\mathbf{x}, \mathbf{u}) = \mathbf{A}\mathbf{x} + \mathbf{B}\mathbf{u}, \quad (2)$$

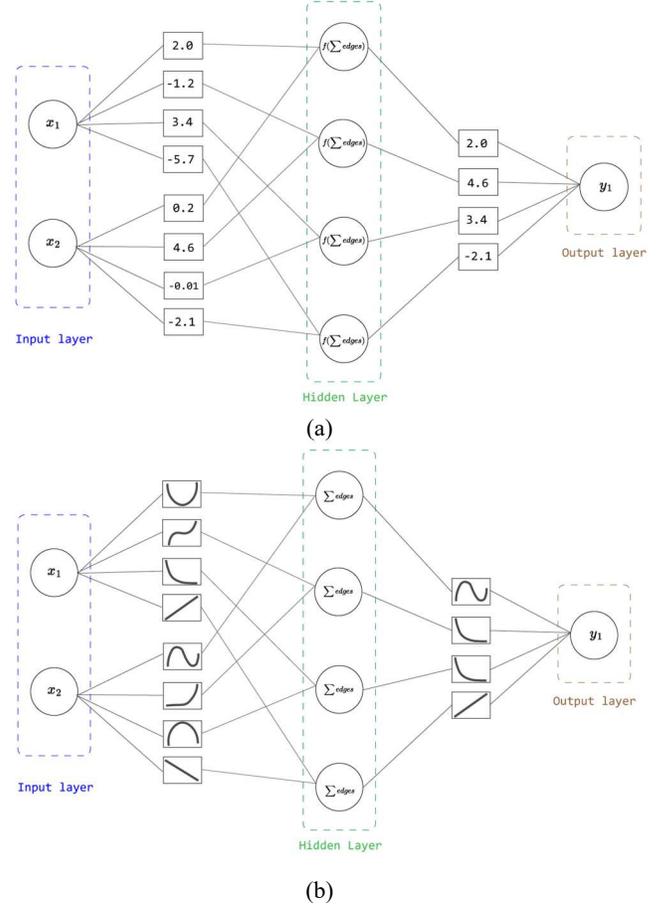

Fig. 1. Network structure in MLPs (a) and KANs (b).

where $\mathbf{x} \in R^n$ represents the system states (e.g., voltage, current, speed, or torque), and $\mathbf{u} \in R^n$ denotes the inputs (e.g., input voltage or duty cycle). Although the assumption here is that the number of inputs matches the number of states, this may not always hold true for every system. By leveraging the inputs and states over time, the derivatives of the states can be approximated using KAN networks:

$$\frac{d\mathbf{x}}{dt} = KAN(\mathbf{x}, \mathbf{u}, \boldsymbol{\theta}\} \approx f(\mathbf{x}, \mathbf{u}) \approx \hat{\mathbf{x}}, \quad (3)$$

where $\boldsymbol{\theta}$ represents the KAN network parameters. From this point onward, $\hat{x}$ will denote the approximated derivative of the state $x$. To model all relevant states, a state-space matrix can be constructed as follows:

$$\frac{d}{dt}\begin{pmatrix} x_1 \\ x_2 \\ \vdots \\ x_n \end{pmatrix} = \begin{bmatrix} KAN_{x_1}(\mathbf{x}, \mathbf{u}, \boldsymbol{\theta}\} \\ KAN_{x_2}(\mathbf{x}, \mathbf{u}, \boldsymbol{\theta}\} \\ \vdots \\ KAN_{x_n}(\mathbf{x}, \mathbf{u}, \boldsymbol{\theta}\} \end{bmatrix}. \quad (4)$$

This setup implies that a separate KAN network is constructed for each state, with the same inputs $(\mathbf{x}, \mathbf{u})$ provided to all networks, while the target output varies depending on the state being modeled. A visual depiction of this is shown in Fig. 2.

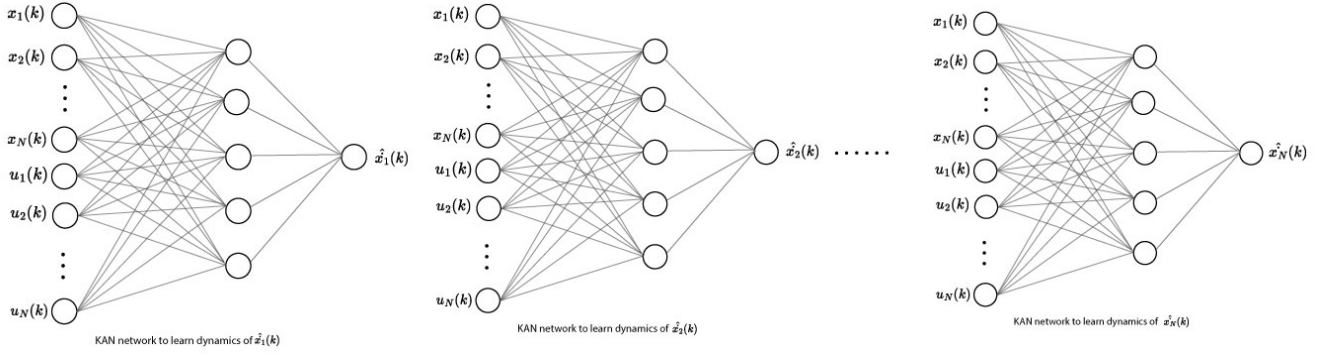

*Fig. 2. Illustration of how KANs learn dynamics.)*

The derivative of each state is computed numerically using finite-difference methods:

$$\hat{\dot{x}}_k = \begin{cases} x_1 - x_0 & \text{for } k = 0 \\ x_{k+1} - x_{k-1}, & \text{for } 1 < k < K - 1, \\ x_K - x_{K-1} & \text{for } k = K \end{cases} \quad (5)$$

where $K$ is the total amount of samples. Note that we do not divide by $\Delta t$ since then the derivative values will become very large and the network is found to have issues with learning, but rather the factor of $\Delta t$ is compensated at the final stage. If the KAN network successfully approximates $f(x, u)$, and if $f(x, u)$ is a smooth function, the composition of the KAN network allows the discovery of the underlying equation for $f(x, u)$ [1]:

$$f(x, u) = (\Phi_{L-1} \circ \Phi_L \circ \ldots \circ \Phi_1 \circ \Phi_0) g, \quad (6)$$

where $\Phi_1$ represents the learned activation functions at each layer, and $g = [x\ u]$, corresponding to the system's dynamic states and inputs. This methodology is inspired by prior work [11] as well as residual networks [12].

### III. APPLICATION TO BUCK CONVERTERS

A buck converter with voltage control is considered as an example to demonstrate the development and application of KANs. In particular, the pulse-width modulation (PWM) uses a triangular carrier signal, with the inductor current and output voltage (system states) sampled at the carrier triangle's peaks. This sampling strategy is widely adopted in power electronics as it reduces noise caused by switching events and captures the average values of the states. Since the average values of all system states and inputs are sampled, higher-order effects, such as switching ripples, are neglected. As a result, the system is effectively treated as linear.

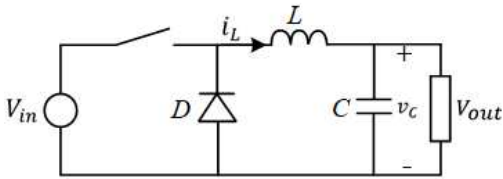

*Fig. 3. Schematic of a buck converter.*

The system under consideration has two states: the inductor current ($i_L$) and the capacitor voltage ($v_C$), where the capacitor voltage is equivalent to the output voltage ($v_C = v_{out}$). Additionally, parasitic resistances on the capacitor and inductor, as well as the on-resistance of the switches are considered. Finally, the system includes two inputs: the duty cycle ($D$) and the input voltage ($V_{in}$). Thus, the states and inputs are represented as $x = [i_L, v_C]$ and $u = [D, V_{in}]^T$, respectively. To model the dynamics, two separate KAN networks are developed and trained. One network approximates $\hat{i}_L \approx \frac{di_L}{dt}$ and the other approximates $\hat{v}_C \approx \frac{dv_C}{dt}$. Both networks share the same inputs (states and control inputs), but the label differs, with one focusing on the current derivative and the other on the voltage derivative.

For simplicity, the input voltage ($V_{in}$) is kept constant during the simulation to reduce computational complexity. As a result, the input vector to the KAN networks is defined as $g = [i_L\ v_C\ D]$. The inputs and labels are treated as discrete-time data sampled at twice the switching frequency ($T_s = \frac{1}{2f_{sw}}$). Throughout the dataset, $k$ denotes the current sample index, while $K$ represents the total number of samples. The parameters of the investigated buck converter are $R = 3\Omega$, $C = 720\mu F$, and $L = 10.3\mu H$. The applied switching frequency was $f_{sw} = 20$kHz. With the inputs to the KAN network properly defined, the next step is to prepare the dataset required for training. The simulation lasts about 5 seconds, with changing voltage output reference to include some transient conditions, as shown in Fig. 4. It is worth noting that if considering a longer training dataset enriched with more variations, the performance of the approach would be expected to improve significantly, however data collection is not always easy thus a more realistic case was considered.

Firstly, the KAN model is trained to predict voltage dynamics, specifically $\frac{dV_{out}}{dt}$. Before proceeding, it is necessary to adjust the grid of the splines, ensuring they align with the range of the input data. The next step involves defining a custom loss function for the training process. The creator of this workflow has observed that when dealing with "small values" in the range of $10^{-1}$ or smaller, the *log-cosh* loss function tends to perform better than the standard mean-squared error (MSE) loss function. This observation is relevant here, as the derivative values we aim to fit fall within this range. Notably, these small values arise because the numerical derivative is not divided by $\Delta t = T_s$. With this in mind, we will implement the *log-cosh* loss function as follows:

$$\mathcal{L}(\theta, g, y_t) = \sum_{k}^{K} \log(\cosh(y_{predicted}(\theta, g) - y_{true})), \quad (7)$$

where $\theta$ are the model parameters, and $y_{true} = \hat{x}$ is the numerical derivative. *Log-cosh* is expected to perform satisfactorily, as the loss is quadratic for small values and linear for large values. Moreover, the loss function includes also a regularization term introduced in [10] to encourage the network to become sparser and prevent overfitting.

To train the voltage dynamics, the hyperparameters were selected through a combination of manual tuning and heuristic approaches. While effective for this case, it is worth observing that a more automated method, such as grid search or Bayesian optimization, could be implemented to streamline and optimize the process further. After experimentation, the following hyperparameters were determined to yield satisfactory results: *opt*=LBFGS, *steps*=60, *loss_fn*=loss_logcosh, *lamb*=0.01, and *lamb_entropy*=10. These values were found to balance accuracy and computational efficiency, ensuring the model adequately fit the data while maintaining smoothness and sparsity in the learned representation. The LBFGS optimizer was found more stable with good convergence behavior compared to the most commonly used ADAMs [10].

## IV. RESULTS AND DISCUSSION

In this section, the performance of the proposed approach is evaluated through various analyses, including system identification and model verification. The results demonstrate how KANs approximate system dynamics and provide insights into the underlying mathematical structure.

### A. System identification

This subsection details the outcomes of training the KAN models, emphasizing their capacity to accurately estimate state derivatives and represent the system's dynamics.

#### 1) Voltage dynamics

Upon completing the training, the learned activation functions of all edges, depicted in Fig. 5, provide valuable insights. It is evident that the majority of these functions exhibit a linear behavior, aligning with the expected dynamics of the buck converter. Minor deviations from linearity, attributed to ripple noise, are also observed but remain minimal, further validating the model's accuracy in capturing the underlying system behavior. The true and predicted derivatives from the trained KAN network with the same input on which it was trained are shown in Fig. 6, exhibiting high accuracy in following the system response.

After verifying that the KAN has successfully learned and that the network primarily consists of linear functions, we proceed to convert the learned splines into explicit mathematical expressions using symbolic regression. KAN includes a function, *fix_symbolic()*, which replaces the learned activation functions with precise symbolic representations, allowing us to extract interpretable equations from the network. However, the third activation function, corresponding to the duty-cycle input, appears somewhat ambiguous. As a result, we will refrain from converting it to a symbolic form, focusing instead on the other functions that clearly exhibit linear behavior.

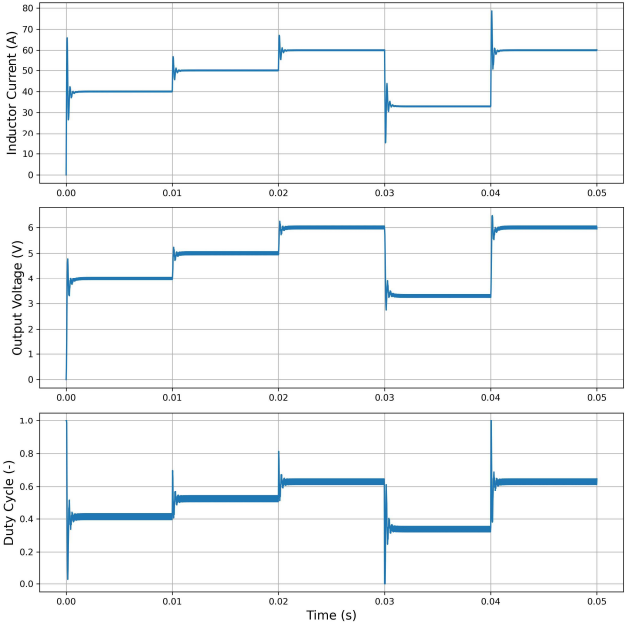

(a)

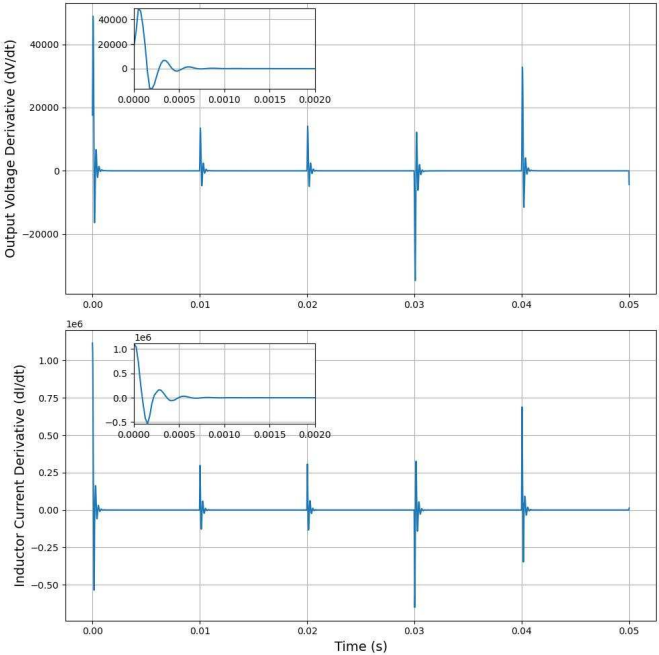

(b)

*Fig. 4*. Overview of the training set. (a) The measured states and input duty cycle. (b) The numerical derivative of the inductor current and output voltage.

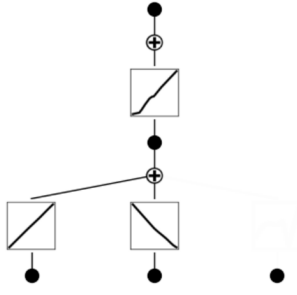

Fig. 5. KAN structure after the first training round.

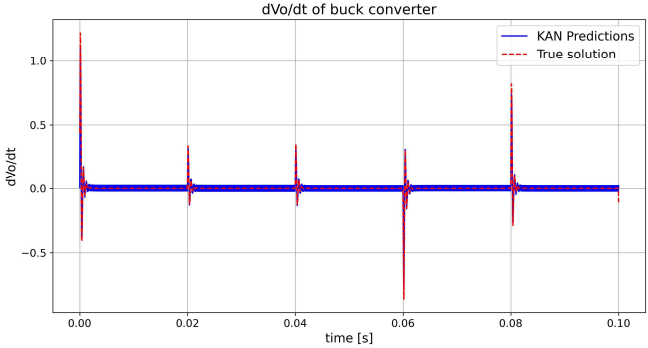

(a)

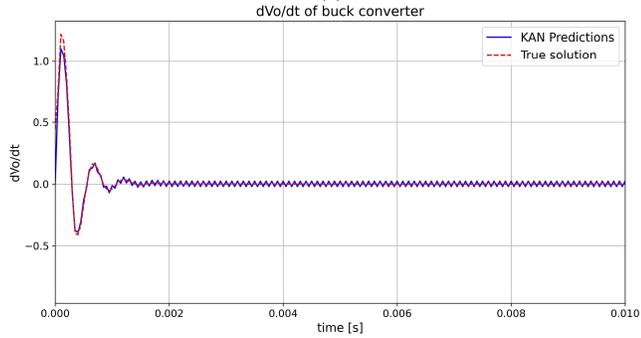

(b)

Fig. 6. Dynamics of the capacitor voltage for true and predicted derivatives from KAN.

Now, they are all fixed to linear, and the color turns red to indicate that the activations are fixed with specific functions, as shown in Fig. 7. We can see that the third input to the model (duty cycle in our case) is completely fainted away, this indicates that duty cycle has no impact on the output data. So now we can fix the third input to simply zero to discard it. The system equation and parameters can now be retrieved by KAN as follows:

$$\frac{dv_c}{dt} = 1179.94 i_L - 11840 v_c + 160. \quad (8)$$

We can conclude that the equation above is the learned $\frac{dV_{out}}{dt}$. It is important to multiply the result with $\frac{1}{T_s}$ to remove the $T_s$ factor from the numerical derivative.

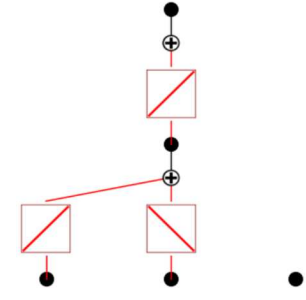

Fig. 7. KAN structure after fixing selected functions.

*2) Current dynamics*

After discovering the voltage dynamics, the current dynamics are identified by following the same process. In the first training round, KAN returns the system shown in Fig. 8. The result of the true and predicted system behaviors is satisfactory, as shown in Fig. 9. The current dynamics equation is as follows:

$$\frac{di_L}{dt} = -2871.79 i_L - 70704 v_c + 917064 D + 19130.43. \quad (9)$$

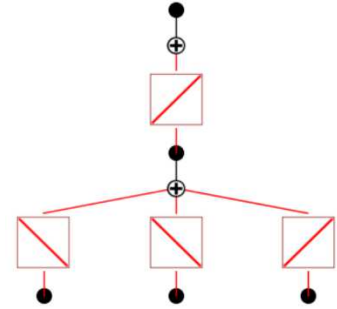

Fig. 8. KAN structure after fixing selected functions.

### B. Model verification

Here the identified models are validated by comparing their predictions with ground truth data, assessing their accuracy and generalization capabilities. More specifically, since the equations of the system have been identified, the state-space matrix can be constructed and used to compare the results of the predicted system against the actual one. Since the input voltage has not been introduced, KAN has added some "discrepancy" to compensate it. To include those "discrepancies", the system inputs must be augmented to contain the duty cycle $D$ and a variable $\gamma$ which will be constant and equal to 1. The state space equations are then:

$$\hat{\dot{x}} = Ax + Bu, \quad (10)$$

$$\begin{bmatrix} \hat{i}_L \\ \hat{v}_C \end{bmatrix} = \begin{bmatrix} -2871.79 & -70704 \\ 1179.94 & -11804.04 \end{bmatrix} \begin{bmatrix} i_L \\ v_C \end{bmatrix} + \begin{bmatrix} 917064 & 19130.43 \\ 0 & 160 \end{bmatrix} \begin{bmatrix} D \\ \gamma \end{bmatrix}, \quad (11)$$

and the output is then given as below:

$$y = Cx + Du, \quad (12)$$

$$y = [0 \quad 1]x. \quad (13)$$

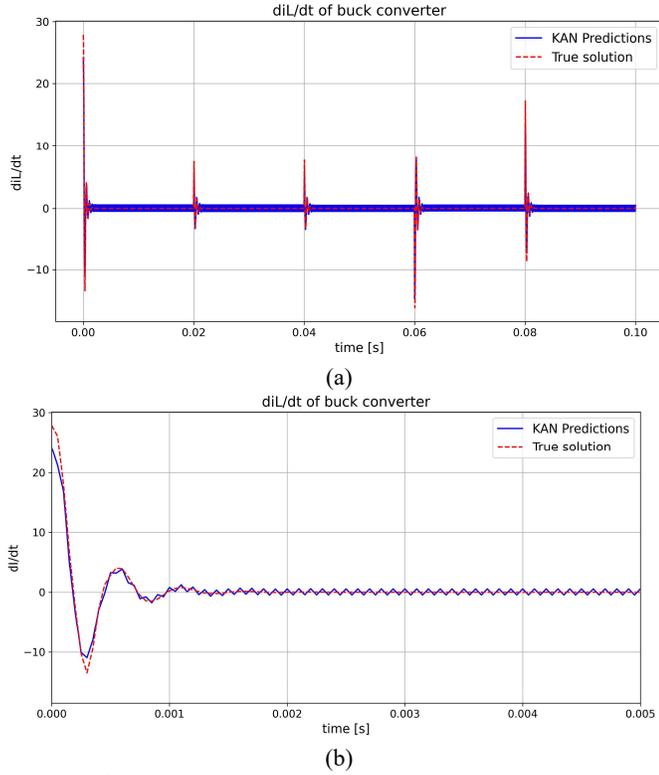

Fig. 9. Dynamics of the inductor current for true and predicted derivatives from KAN.

In Fig. 10, it is evident that the KAN solution and the true buck-converter dynamics are in good agreement. Thus, it can be safely concluded that the dynamins of the buck-converter average state-space model has been learned by KAN only from the data as a source input. KAN managed not only to discover the system equations, but also to estimate its parameters. To improve the accuracy, more training data can be used with fixed symbolic functions. Additionally, the input voltage, which was considered constant, can be changed to understand how it affects the system as well.

## V. CONCLUSION

This study demonstrated the effectiveness of KANs for system identification and model verification in dynamic systems, with a focus on a buck converter. KAN successfully approximated state derivatives and captured system behavior through interpretable activation functions, which were further refined into symbolic representations using regression techniques. This process highlighted KAN's ability to discover the underlying linear dynamics of the system while maintaining high accuracy and interpretability. Additionally, the network proved robust in identifying system parameter changes, showcasing its potential for real-time industrial applications.

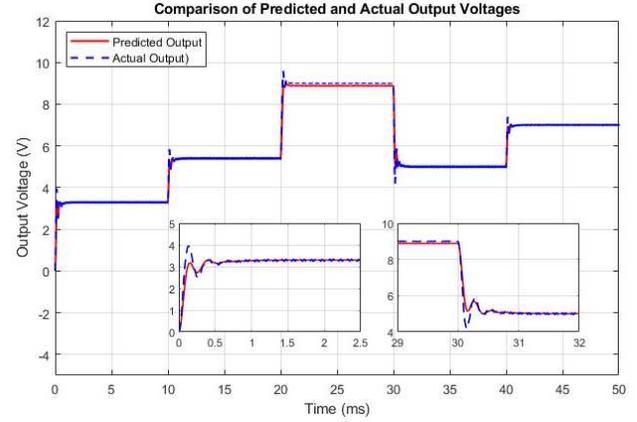

Fig. 10. Comparison of the output voltage from the simulation model and the estimated system from KAN.